# STDP Based Pruning of Connections and Weight Quantization in Spiking Neural Networks for Energy-Efficient Recognition

Nitin Rathi, Priyadarshini Panda, *Student Member, IEEE*, and Kaushik Roy, *Fellow, IEEE*

*Abstract*— Spiking Neural Networks (SNNs) with a large number of weights and varied weight distribution can be difficult to implement in emerging in-memory computing hardware due to the limitations on crossbar size (implementing dot product), the constrained number of conductance levels in non-CMOS devices and the power budget. We present a sparse SNN topology where non-critical connections are pruned to reduce the network size and the remaining critical synapses are weight quantized to accommodate for limited conductance levels. Pruning is based on the power law weight-dependent Spike Timing Dependent Plasticity (STDP) model; synapses between pre- and post-neuron with high spike correlation are retained, whereas synapses with low correlation or uncorrelated spiking activity are pruned. The weights of the retained connections are quantized to the available number of conductance levels. The process of pruning non-critical connections and quantizing the weights of critical synapses is performed at regular intervals during training. We evaluated our sparse and quantized network on MNIST dataset and on a subset of images from Caltech-101 dataset. The compressed topology achieved a classification accuracy of 90.1% (91.6%) on the MNIST (Caltech-101) dataset with 3.1x (2.2x) and 4x (2.6x) improvement in energy and area, respectively. The compressed topology is energy and area efficient while maintaining the same classification accuracy of a 2-layer fully connected SNN topology.

1. ## INTRODUCTION

Human brain consisting of 20 billion neurons and 200 trillion synapses is by far the most energy-efficient neuromorphic system with cognitive intelligence. The human brain consumes only ~20W of power which is nine orders of magnitude lower compared to a computer simulating human brain activity in real time [1]. This had led to the inspiration and development of Spiking Neural Networks (SNNs) which tries to mimic the behavior of human brain and process inputs in real time [2]. SNNs may provide an energy-efficient solution to perform neural computing. However, recent works have shown that to get reasonable accuracy compared to non-spiking Artificial Neural Networks (nANNs), the complexity and size of SNNs is enormous. In [3] to improve the classification accuracy for MNIST dataset by 12%, the number of neurons in 2-layer SNN had to be increased by 64x. The authors in [4] achieved an average accuracy of 98.6% for MNIST dataset with two hidden layers consisting of 800 neurons in each layer. The quest of making SNNs larger and deeper for higher accuracy have compromised their energy efficiency and introduced challenges as mentioned below:

1. Large SNNs implemented on emerging memristive crossbar structures [5] are limited by the crossbar size. Large crossbars suffer from supply voltage degradation, noise generated from process variations, and sneak paths [6, 7].

2. SNNs with numerous synapses involve higher number of computations making them slower and energy inefficient.

SNNs are driven by the synaptic events and the total computation, memory, communication, power, area, and speed scale with the number of synapses [2]. We propose a pruned and weight quantized SNN topology with self-taught Spike Timing Dependent Plasticity (STDP) based learning. STDP, in turn, is also used to classify synapses as critical and non-critical. The non-critical synapses are pruned from the network, whereas the critical synapses are retained and weight quantized. Such pruning of connections and weight quantization can lead to their efficient implementations in emerging cross-bar arrays such as resistive random access memories (R-RAMs) [8, 9], magnetic tunnel junctions [10], or domain-wall motion based magnetic devices [11]. Such cross-bars, even though suitable for implementing efficient dot-products required for neural computing, are constrained in size, because of non-idealities such as sneak paths, weight quantization, and parameter variations [6, 7]. The resulting sparse SNN can achieve 2-3x improvement in energy, 2-4x in area and 2-3x in testing speed.

Synaptic pruning is commonly observed during the development of human brain. The elimination of synapses begins at the age of two and continues till adulthood, when synaptic density stabilizes and is maintained until old age [12]. From hardware implementation of neural networks, synapses are a costly resource and needs to be efficiently utilized for energy efficient learning. If synapses or connections are properly pruned, the performance decrease due to synaptic deletion is small compared to the energy savings [13]. This has motivated researchers to apply the technique of pruning [14] and weight quantization [15] to compress nANNs. Pruning and quantization performed on state-of-the-art network AlexNet trained for ImageNet dataset provided 7x benefit in energy efficiency along with 35x reduction in synaptic weight storage without any loss of accuracy [15]. The authors in [14] prune the connections of an nANN trained using backpropagation based on the Hessian of the loss function. The number of parameters were reduced by a factor of two while maintaining the same test accuracy. The supervised learning algorithm in [16] pruned the hidden layer neurons with low dominance to reduce network size. The network achieved similar performance with 4x less parameters for Fisher Iris problem compared to other spiking networks. The idea of pruning is based on identifying parameters with small saliency, whose deletion will have minimal effect on the error. These networks were trained with supervised learning algorithms, but in SNNs with unsupervised training it is difficult to calculate such parameters since there is no such defined error function. In real world, obtaining unlabeled images for unsupervised learning is much easier than gathering labelled images for supervised learning. The novelty of our approach lies in self-taught STDP based weight pruning where the connections to be pruned are decided based on their weights learnt by the unsupervised STDP algorithm. Connections having STDP weights above a threshold are considered critical while others are temporarily pruned. The threshold is fixed before training and it referred as pruning threshold. The critical connections are weight quantized to further reduce network complexity. The resulting compressed topology is energy-efficient while maintaining accuracy and alleviates the issues that constrain the scalability of crossbar structure, leading to robust design of neuromorphic systems.

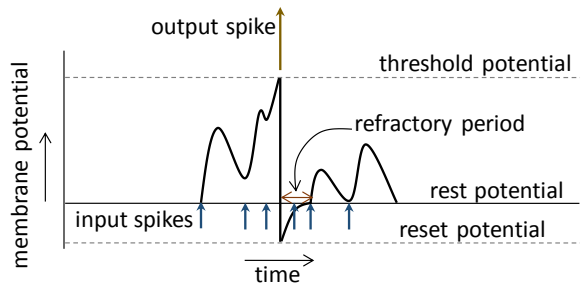

Fig. 1. Leaky-Integrate-and-Fire (LIF) model of a single neuron's membrane potential dynamics in response to input spikes in SNN.

The rest of the paper is organized as follows. Section 2 provides background information on the neuron and the synapse models and the STDP learning algorithm employed in this work. The network topology and the training and testing schemes are also briefly discussed. Section 3 presents the proposed compression techniques; STDP based pruning and weight quantization and sharing. The experiments on the proposed topology are presented in Section 4. The results of the experiments are analyzed in Section 5. Conclusions are drawn in Section 6.

## 2. BACKGROUND

### 2.1. Neuron & Synapse Model and STDP Learning

We employ the Leaky-Integrate-and-Fire (LIF) model [3] to simulate the membrane potential dynamics of a neuron in our spiking network model. Fig. 1 shows the change in membrane potential of a single post-neuron in response to input spikes (blue arrows) from pre-neurons. The membrane potential increases at the onset of a spike and exponentially decays towards rest potential in the absence of spiking activity. The post-neuron fires or emits a spike when its potential crosses the threshold and immediately its potential is set to a reset value. After firing the post-neuron goes into a period of inactivity known as refractory period during which it is abstained from spiking, irrespective of input activity as shown in Fig. 1.

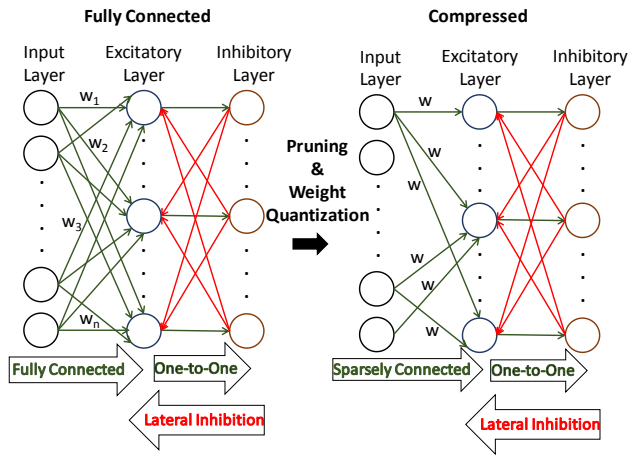

Fig. 2 SNN topology with lateral inhibition. Input to excitatory is fully connected which is later pruned. Excitatory to inhibitory is one-to-one connected, whereas inhibitory is backward connected to all the excitatory except the one it receives the connection from. Pruning is performed only on the input to excitatory connections.

The connection between two neurons is termed a synapse and is modelled by the conductance change which is modulated by the synaptic weight (*w*). The synaptic weight between a pair of neurons increases (decreases) if the post-neuron fires after (before) the pre-neuron has fired. This phenomenon of synaptic plasticity where the weight change is dependent on the inter spike timing of pre- and post-neuron is termed STDP. We adopt the power law weight-dependent STDP model, where the weight change is exponentially dependent on the spike timing difference of the pre- and post-neuron ($t_{pre} - t_{post}$) as well as the previous weight value [3].

### 2.2. Network Topology

The SNN topology for this work is shown in Fig. 2. It consists of input layer followed by excitatory and inhibitory layer. The input layer is fully connected to the excitatory layer, which in turn is one-to-one connected to the inhibitory layer. The number of neurons in the excitatory layer are varied to achieve better accuracy, whereas the number of neurons in the inhibitory layer is the same as the number in the excitatory layer. Each inhibitory neuron is backward connected to all the excitatory neurons except for the one from which it receives a connection from. Thus, the inhibitory layer provides lateral inhibition which discourages simultaneous firing of multiple excitatory neurons and promotes competition among them to learn different input features. The process of pruning and weight quantization is applied to the excitatory synapses to obtain the compressed topology as shown in Fig. 2. The fully connected topology serves as the baseline design and we compare the results for the compressed design with baseline.

To ensure similar firing rates for all neurons in the excitatory layer we employ an adaptive membrane threshold mechanism called homoeostasis [3]. The threshold potential is expressed as $V_{thresh} = V_t + \theta$, where $V_t$ is a constant and θ is changed dynamically. θ increases every time a neuron fires and decays exponentially. If a neuron fires more often, then its threshold potential increases and it requires more inputs to fire again. This ensures that all neurons in the excitatory layer learn unique features and avoids few neurons from dominating the response pattern.

### 2.3. Training & Testing

The connections from input to excitatory layer are trained using the STDP weight update rule to classify an input pattern. The training is unsupervised as we do not use any labels to update the weights. The weight update is given by the formula described in section 3.1. The input image is converted into a Poisson spike train based on individual pixel intensities. The excitatory neurons are assigned a class/label based on their average spiking activity over all the images. During testing, the class prediction is inferred by averaging the response of all excitatory neurons per input. The class represented by the neurons with the highest spiking rate is predicted as the image label. The prediction is correct if the actual label matches the one predicted by the SNN. This is similar to the approach followed in [3].

### 3. COMPRESSION TECHNIQUES

In this section, we describe the two compression techniques (pruning and weight quantization) employed in this work to convert the 2-layer fully connected SNN into a compact and sparse topology for digit and image recognition.

### 3.1. STDP Based Pruning

Spike Timing Dependent Plasticity (STDP) is widely used as an unsupervised Hebbian training algorithm for SNNs. STDP postulates that the strength of the synapse is dependent on the spike timing difference of the pre- and post-neuron. The power law weight update for an individual synapse is calculated as

$$\Delta w = \eta \times [e^{\left(\frac{t_{pre}-t_{post}}{\tau}\right)} - offset] \times [w_{max} - w]^\mu$$

where $\Delta w$ is the change in weight, $\eta$ is the learning rate, $t_{pre}$ and $t_{post}$ are the time instant of pre- and post-synaptic spikes, $\tau$ is the time constant, $offset$ is a constant used for depression, $w_{max}$ is the maximum constrained imposed on the synaptic weight, $w$ is the previous weight value, $\mu$ is a constant which governs the exponential dependence on previous weight value. The weight update is positive (potentiation) if the post-neuron spikes immediately after the pre-neuron and negative (depression) if the spikes are far apart (Fig. 3). We employ STDP to train the excitatory synapses as well as to classify them as critical or non-critical. The synapses whose weights do not increase for a set of inputs are likely to have not contributed towards learning and thus can be potential candidates for deletion. On the other hand, synapses with higher weights have most likely learned the input pattern and can be classified as critical (provided they were initialized with small weights). The characteristic features of the input is captured in connections with higher weights and are critical for correct classification. Thus, synapses with STDP trained weights ( $w + \Delta w$) above pruning threshold are considered critical and all other synapses are marked as non-critical. The process of pruning and training is performed repeatedly by dividing the entire training set into multiple batches. After each batch the weights of all non-critical synapses are reduced to zero (they still remain in the network) and the network is trained with the next batch. The synapses with zero weight continue to participate in training, thus a non-critical synapse may become critical for different inputs. The process of reducing the weight to zero instead of eliminating the connection is essential for the network to learn the representation of inputs which appear in latter batches. The elimination of synapses will either make the network not learn the new representations or force the network to forget previous representations in order to learn new inputs. The process of retaining non-critical synapses with zero weight makes the network scalable. In the final training step when all the training images have been presented to the network any remaining non-critical connections are permanently removed from the network. The training starts with a fully connected network and the number of critical connections gradually decrease over time. At the end of training only the critical connections capturing the characteristic features of the inputs remain.

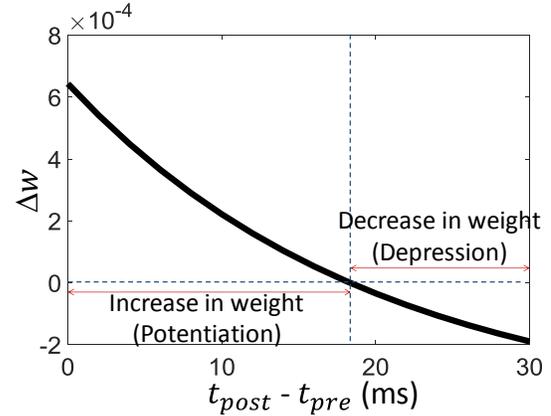

Fig. 3 Change in synaptic weight based on temporal correlation in pre- and post-synaptic spikes. ($\eta = 0.002, \tau = 20$ ms, $offset = 0.4, w_{max} = 1, w = 0.5, \mu = 0.9$)

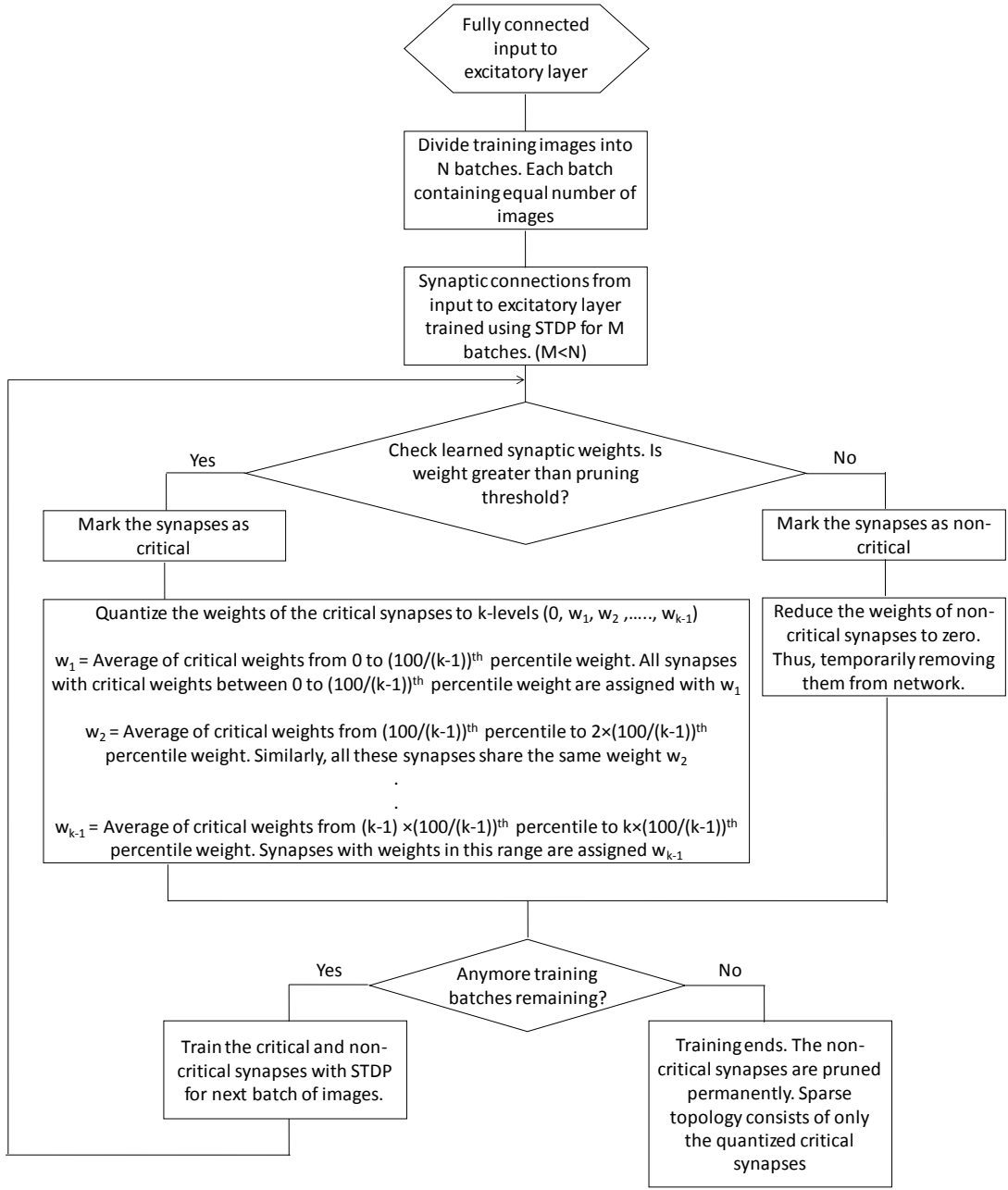

Fig. 4 Flowchart of the proposed algorithm for compressing SNN using pruning and weight quantization.

## 3.2. Weight Sharing and Quantization

The process of pruning reduces the overall connectivity, but as mentioned in section 1, SNNs with continuous weight values are difficult to implement in crossbar structures due to limitations on the number of available conductance levels in devices implementing the synapse. Weight sharing and quantization discretizes the weights to the available number of conductance levels. For example, network with 2-level weight quantization has only two values of weights: 0 (no connection) and $w$. All the synapses share the same weight ($w$) and the entire network can be represented as a sparse binary matrix. A 2-level weight quantized SNN can be implemented in crossbar architecture with a single fixed resistor [17], where $w$ is the conductance of the resistor. The value of $w$ is the average weight of all the critical connections trained using STDP. For example, we start with a network with n number of synapses and after training (with pruning) m critical synapses remain. The weights of the m critical synapses ($w_1$, $w_2$, ….., $w_m$) are continuous and computed based on the STDP formula. The common weight value $w$ is the average of $w_1$ to $w_m$. The average is calculated after each pruning step and all the critical connections share the same average weight ($w_1$ to $w_m$ is replaced with $w$). Like pruning, the process of weight quantization and sharing is performed repeatedly after each training batch. The value of $w$ changes at every quantization step and the final value is obtained after training the network for all the input batches. Similarly, the weights can be quantized to 3-levels: 0, $w_1$, $w_2$, where $w_1$ ($w_2$) is the low (high) conductance value. The conductance values are computed by calculating the 50$^{th}$ percentile or the median weight of all the critical connections. The lower conductance value $w_1$ is the average of all weights between 0 and the median weight, $w_2$ is the average of rest of the weights. The critical synapses with weights between 0 and the median weight are assigned $w_1$. The critical synapses with weights between median weight and the maximum weight share the quantized vale of $w_2$. The accuracy of the network is directly proportional to the number of quantization levels. The performance of the system improves with more number of conductance levels. In a quantized SNN most of the connections share the same weight which reduces the implementation complexity.

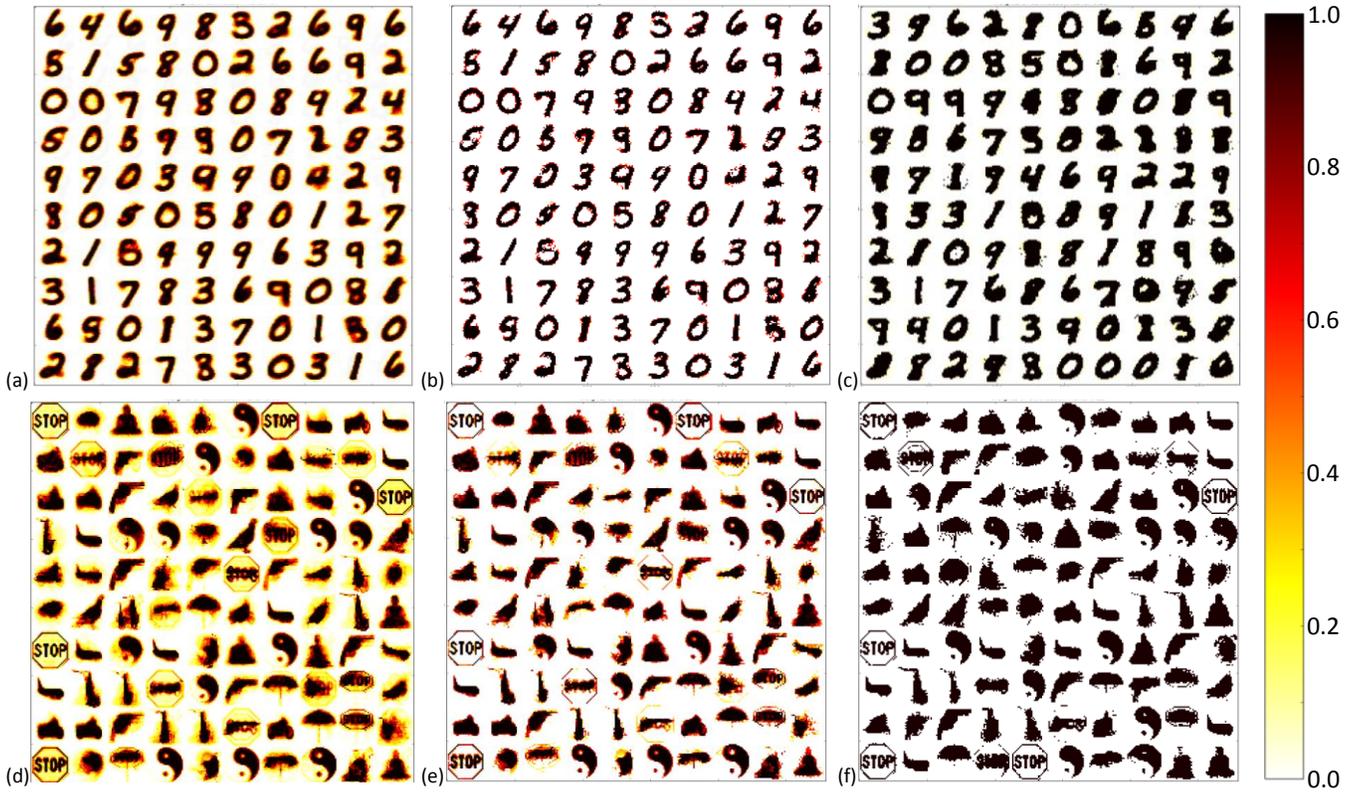

Fig. 5 Rearranged weights of the connections from input to excitatory for (a) MNIST baseline; (b) MNIST pruning; (c) MNIST pruning and quantization; (d) Caltech 101 baseline; (e) Caltech 101 pruning; (f) Caltech 101 pruning and quantization

Fig. 4 summarizes the proposed algorithm for achieving a pruned and weight quantized SNN. The 2-layer untrained network is initialized with full connectivity from input to excitatory layer. The weights are randomly assigned from a uniform distribution. The training images are divided into N batches of equal number of images. The excitatory synapses are trained with STDP weight update rule for M (M<N) training batches. The connections with current weights above the pruning threshold are classified as critical, rest of the connections are marked as non-critical. The non-critical connections are pruned by reducing their weights to zero. The weights of the critical synapses are quantized to the required number of conductance levels. The pruned and quantized network is trained with STDP weight update rule for the next training batch. The process of pruning and quantization is performed at regular intervals for all the remaining training batches. The training ends when all the batches have been presented to the network. The first pruning and quantization step is delayed for M batches to ensure proper detection of critical connections and to mitigate the bias due to random initialization of weights. The randomly initialized synapses require more training images to capture the input characteristic features. Once the input features have been captured the pruning can be performed more often (after every batch). Once the critical connections are identified, they more or less remain the same during training. So, the first pruning step is very crucial and more than one training batch is needed to identify the critical synapses. The baseline design is trained in a similar fashion with no pruning and quantization. All the training images are presented in one batch and the weights are trained using STDP.

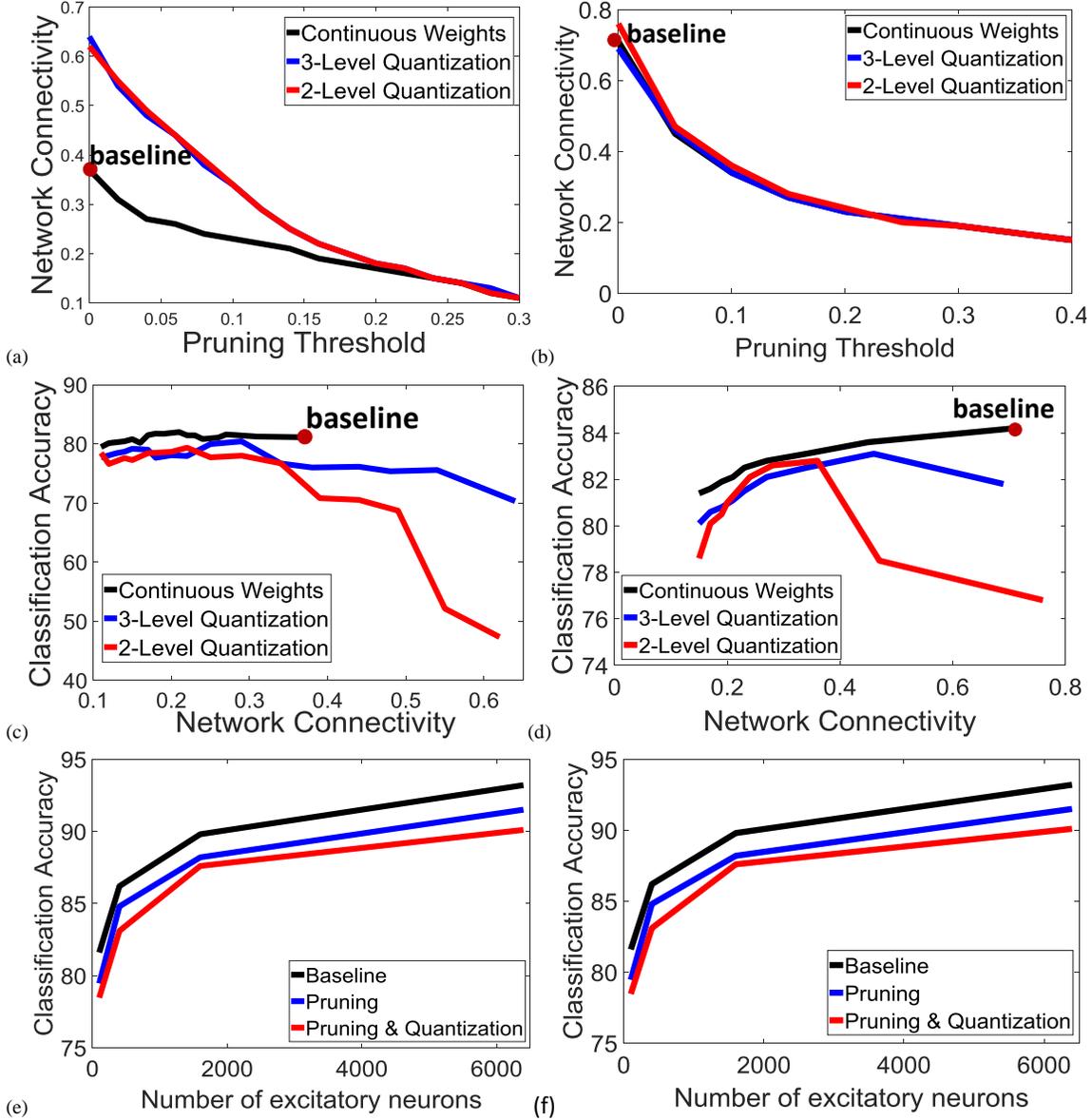

Fig. 6 Variation in network connectivity with pruning threshold for (a) MNIST; (b) Caltech 101. Classification accuracy for different network connectivity for (c) MNIST; (d)Caltech 101. Classification accuracy for different number of excitatory neurons for (e) MNIST; and (f) Caltech 101.

## 4. EXPERIMENTAL METHODOLOGY

The proposed SNN topology is simulated in the open source spiking neuron simulator BRIAN implemented in Python [18]. BRIAN allows the modelling of biologically plausible neurons and synapses defined by differential equations. The parameters for the models are same as [3]. We tested our network for digit recognition on the MNIST dataset [19] and image recognition on a subset of images from the Caltech 101 dataset [20].

### A. MNIST Dataset

MNIST dataset contains 28×28-pixel sized grayscale images of digits 0-9. Thus, the input layer has 784 (28×28) neurons fully connected with 100 excitatory neurons. The dataset is divided into 60,000 training and 10,000 testing images. We further divide the 60,000 training images into batches of 5,000 images (N=12). The baseline design is trained with entire 60,000 images presented one after another. The compressed topology is initially trained for three training batches totaling 15,000 images (M=3). STDP based critical connections are weight quantized and the weights of the non-critical connections is reduced to zero. The pruned and quantized network is trained with the next training batch. The process of pruning and quantization is performed after every batch henceforth. The rearranged input to excitatory synaptic weights of the trained baseline topology with 100 excitatory neurons is shown in Fig. 5(a). Fig. 5(b) shows the rearranged synaptic weights of the same network compressed with pruning threshold=0.3 and having continuous weight values. The rearranged synaptic weights of the pruned and 2-level weight quantized network is shown in Fig. 5(c).

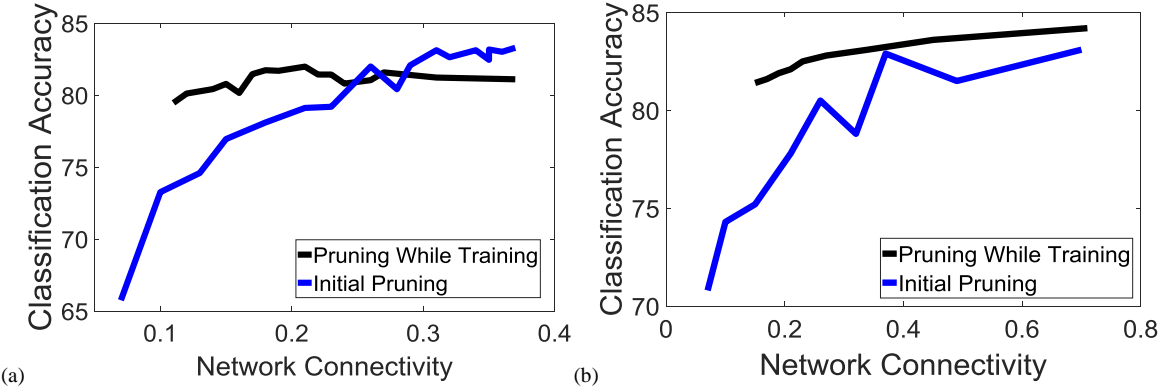

Fig. 7 Classification accuracy for different network sparsity achieved by pruning the connections during training and before training for (a) MNIST; and (b) Caltech 101.

## B. Caltech 101 Dataset

Caltech 101 dataset is a collection of images of objects belonging to 101 different categories. Each category consists of 40 to 800 of around 300×200-pixel sized RGB images. The dataset also provides annotations for the object in the image which we use to separate the object from the background. Unlike MNIST images, we preprocess the Caltech 101 images to obtain 28×28-pixel sized grayscale images. Maintaining the same image size across datasets ensures that we do not need to change the network parameters. Out of 101, we selected 10 categories (yin yang, saxophone, stop sign, wrench, revolver, Buddha, airplanes, pigeon, motorbikes, umbrella) and randomly divided the total images in each category with 80% training and 20% testing images. Since each category has different number of images we create copies of images so that each category has similar number of training and testing images. This is necessary to avoid categories with more images to dominate the learning in the network. The preprocessing steps involved converting the images to grayscale, averaging the pixels with Gaussian kernel of size 3×3 to suppress the noise and resizing the image to 28×28 pixels. All the preprocessing steps are performed using the OpenCV library [21] in python. The training set consists of 10,000 images with 1000 images per category. The 10,000 images are further divided into batches of 500 images (N=20). The baseline fully connected design is trained with entire 10,000 images. The compressed topology is initially trained with ten training batches totaling 5000 images (M=10). The critical connections are identified using STDP and weight quantized. The non-critical synapses are pruned. The pruned and quantized network is trained with all the remaining batches with pruning and quantization performed after every training batch. Fig. 5(d), (e) and (f) show the rearranged synaptic weights for the baseline, pruned and weight quantized topologies, respectively. Compression is performed with pruning threshold of 0.2 and 2-level weight quantization.

## 5. RESULTS & ANALYSES

In this section, we analyze the results and compare the performance of compressed topology with the baseline design. The results are evaluated based on different parameters like pruning threshold and number of excitatory neurons. The removal of connections during training is compared with training a sparse network, both having similar connectivity.

### 5.1. Comparison with varying pruning threshold

The network connectivity is a strong function of the pruning threshold; higher the threshold, sparser is the network. The network connectivity is defined as the ratio of the actual number of connections to the total number of possible connections. The total number of possible connections with 100 excitatory neurons is 78400 (784×100). The number of actual connections depend on the pruning steps. Fig. 6(a) and (b) shows the variation in final network connectivity with pruning threshold for MNIST and Caltech 101 datasets, respectively. The red dot in Fig. 6(a) and (b) with zero pruning threshold denotes the baseline design with no compression techniques applied. Ideally, the connectivity should be 1 since the connections are not pruned during training. The reduction in connectivity results from the inherent depression in the STDP learning rule. The further reduction in connectivity is achieved by increasing the pruning threshold. The compressed topologies are less sparse for low pruning threshold compared to baseline. This is due to weight quantization and sharing in early training stages. The shared weight is the average of all critical weights which is higher than almost half the critical weights. Thus, the average weight replaces half the STDP learnt weights which were supposed to be much lower. This reduces the effect of inherent STDP depression on these synapses and reduces the probability of their removal. Fig. 6(c) and (d) show the test accuracy for different network connectivity for MNIST and Caltech 101 datasets, respectively. The baseline topology has an accuracy of 81.6% (MNIST) and 84.2% (Caltech 101) which is consistent with the results shown in [3]. The highest classification accuracy achieved for the compressed topology is 79.5% (MNIST) and 82.8% (Caltech 101). The accuracy degrades slightly compared to baseline but at the same time there is immense drop in network connectivity. The compressed topology is 75% (36%) sparser than the baseline topology for MNIST (Caltech 101) dataset.

### 5.2. Comparison with varying number of neurons

The change in classification accuracy with the number of excitatory neurons for MNIST and Caltech 101 datasets is shown in Figs. 6(e) and 6(f), respectively. The compressed topology has 27% connectivity for MNIST and 42% for Caltech 101 datasets, also the weights are quantized to 3-levels. These parameters correspond to the best performance of compressed topology (Fig. 6(c-d)). The baseline design with 6400 neurons achieved an accuracy of 93.2% for MNIST and 94.2% for Caltech 101 datasets. The pruned topology achieved an accuracy of 91.5% with 8% connectivity and 92.8% with 12% connectivity for MNIST and Caltech 101 datasets, respectively.

### 5.3. Pruning while training

The objective of pruning is to increase the sparsity in the network. This can be achieved in two ways: removing connections during training of a fully connected network or training a sparse network. The second approach is performed by randomly removing connections from a fully connected topology to produce a sparse network. In first case the connections are removed systematically based on some parameters whereas in the second approach the removal is completely random. Nevertheless, in both the cases the final network connectivity is same. Fig. 7(a) and (b) shows the classification accuracy with varying network connectivity for both the approaches for MNIST and Caltech 101 datasets, respectively. The network with initial pruning is trained similar to baseline with no compression techniques. The pruning while training is the approach followed in rest of the paper, where pruning is performed at regular intervals during training. The results for both the networks are shown for continuous weight distribution. Pruning the connections during training performs better since only the non-critical connections are removed. The network with initial sparsity is constructed by randomly removing connections. This shows that STDP successfully identifies the non-critical connections.

### 5.4. Reduction in spike count or energy

The decrease in connectivity due to pruning leads to reduced spiking activity in the excitatory layer. The active power of a SNN is proportional to the firing activity in the network [2]. Thus, the energy can be quantified as the reduction in spike count of excitatory neurons during testing. Fig. 8 shows the normalized reduction in spiking activity or energy for compressed topology with respect to baseline. The pruned topology shows 3.1x and 2.2x improvement in energy whereas the 2-level weight quantized network achieves 2.4x and 1.92x improvement for MNIST and Caltech101 datasets, respectively. The compressed topology may achieve additional energy benefits from implementation in crossbar structures with low power devices. The emerging post-CMOS devices like MTJ, R-RAM and domain wall motion based devices consume very low power in idle state due to elimination of leakage. But these devices have limited number of programmable conductance levels. The compressed topology quantized to the available number of conductance levels can reap the energy benefits provided by these devices. The baseline design with continuous weight distribution is difficult to implement with these devices.

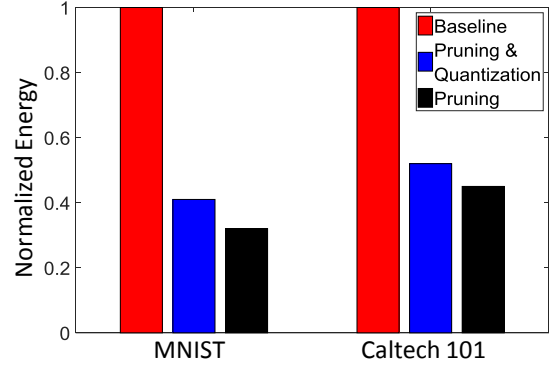

Fig. 8 Normalized improvement in energy with pruning and weight quantization compared to baseline topology.

### 6. CONCLUSIONS

In this paper, we propose two compression techniques, pruning and weight quantization to compress SNNs. Compressed SNNs not only provide energy benefits but also mitigate the issue of limited programmable conductance levels of post-CMOS devices for neuromorphic implementation. The novelty of our approach lies in fact that STDP learning rule is used to decide the network pruning and the weights of the critical connections are quantized to specific levels depending on device and technology requirements. The compressed topology is compared with the 2-layer fully connected topology for digit recognition with MNIST dataset and image recognition with Caltech 101 dataset. The proposed topology achieves 3.1x and 2.2x improvement in energy for MNIST and Caltech 101 datasets, respectively, compared to baseline fully connected SNN. The optimal compression parameters like pruning threshold and weight quantization levels are decided by performing multiple experiments with different images. Additionally, it is worth mentioning that the proposed topology reduced the training time by 3x and 2x for MNIST and Caltech 101 datasets, respectively, by achieving faster training convergence.

ACKNOWLEDGEMENT

The research was funded in part by National Science Foundation, Intel Corporation, Vannevar Bush Faculty

Fellowship, and by the Center for Spintronics (C-SPIN) funded by DARPA/MARCO.

## References

[1] D. S. Modha, "Introducing a brain-inspired computer," Published online at http://www.research.ibm.com/articles/brain-chip.shtml, 2017.

[2] P. A. Merolla, J. V. Arthur, R. Alvarez-Icaza, A. S. Cassidy, J. Sawada, F. Akopyan, B. L. Jackson, N. Imam, C. Guo, Y. Nakamura et al., "A million spiking-neuron integrated circuit with a scalable communication network and interface," Science, vol. 345, no. 6197, pp. 668–673, 2014.

[3] P. U. Diehl and M. Cook, "Unsupervised learning of digit recognition using spike-timing-dependent plasticity," Frontiers in computational neuroscience, vol. 9, 2015.

[4] J. H. Lee, T. Delbruck, and M. Pfeiffer, "Training deep spiking neural networks using backpropagation," Frontiers in neuroscience, vol. 10, 2016.

[5] P. Merolla, J. Arthur, F. Akopyan, N. Imam, R. Manohar, and D. S. Modha, "A digital neurosynaptic core using embedded crossbar memory with 45pj per spike in 45nm," in Custom Integrated Circuits Conference (CICC), 2011 IEEE. IEEE, 2011, pp. 1–4.

[6] B. Liu, W. Wen, Y. Chen, X. Li, C.R. Wu, and T.Y. Ho, "Eda challenges for memristor-crossbar based neuromorphic computing," in Proceedings of the 25th edition on Great Lakes Symposium on VLSI. ACM, 2015, pp. 185–188.

[7] E. Linn, R. Rosezin, C. Kugeler, and R. Waser, "Complementary resistive switches for passive nanocrossbar memories," Nature materials, vol. 9, no. 5, p. 403, 2010.

[8] M. Hu, H. Li, Y. Chen, Q. Wu, G. S. Rose, and R. W. Linderman, "Memristor crossbar-based neuromorphic computing system: A case study," IEEE transactions on neural networks and learning systems, vol. 25, no. 10, pp. 1864–1878, 2014.

[9] S. Park, H. Kim, M. Choo, J. Noh, A. Sheri, S. Jung, K. Seo, J. Park, S. Kim, W. Lee et al., "Rram-based synapse for neuromorphic system with pattern recognition function," in Electron Devices Meeting (IEDM), 2012 IEEE International. IEEE, 2012, pp. 10–2.

[10] P. Krzysteczko, J. Munchenberger, M. Schafers, G. Reiss, and A. Thomas, "The memristive magnetic tunnel junction as a nanoscopic synapse-neuron system," Advanced Materials, vol. 24, no. 6, pp. 762–766, 2012.

[11] M. Sharad, C. Augustine, G. Panagopoulos, and K. Roy, "Spin-based neuron model with domain-wall magnets as synapse," IEEE Transactions on Nanotechnology, vol. 11, no. 4, pp. 843–853, 2012.

[12] P. R. Huttenlocher et al., "Synaptic density in human frontal cortex- developmental changes and effects of aging," Brain Res, vol. 163, no. 2, pp. 195–205, 1979.

[13] G. Chechik, I. Meilijson, and E. Ruppin, "Synaptic pruning in development: a computational account," Neural computation, vol. 10, no. 7, pp. 1759–1777, 1998.

[14] Y. LeCun, J. S. Denker, and S. A. Solla, "Optimal brain damage," in Advances in Neural Information Processing Systems 2, 1990, pp. 598–605. [Online]. Available: http://papers.nips.cc/paper/250-optimalbrain-damage.pdf

[15] S. Han, H. Mao, and W. J. Dally, "Deep compression: Compressing deep neural networks with pruning, trained quantization and huffman coding," arXiv preprint arXiv:1510.00149, 2015.

[16] S. Dora, S. Sundaram, and N. Sundararajan, "A two stage learning algorithm for a growing-pruning spiking neural network for pattern classification problems," in International Joint Conference on Neural Networks (IJCNN), 2015. IEEE, 2015, pp. 1–7.

[17] H. Graf, L. Jackel, R. Howard, B. Straughn, J. Denker, W. Hubbard, D. Tennant, and D. Schwartz, "Vlsi implementation of a neural network memory with several hundreds of neurons," in AIP conference proceedings, vol. 151, no. 1. AIP, 1986, pp. 182–187.

[18] D. Goodman and R. Brette, "Brian: a simulator for spiking neural networks in python," Frontiers in neuroinformatics, vol. 2, 2008.

[19] Y. LeCun, L. Bottou, Y. Bengio, and P. Haffner, "Gradient-based learning applied to document recognition," Proceedings of the IEEE, vol. 86, no. 11, pp. 2278–2324, 1998.

[20] L. Fei-Fei, R. Fergus, and P. Perona, "One-shot learning of object categories," IEEE transactions on pattern analysis and machine intelligence, vol. 28, no. 4, pp. 594–611, 2006.

[21] G. Bradski, "The opencv library." Dr. Dobb's Journal: Software Tools for the Professional Programmer, vol. 25, no. 11, pp. 120–123, 2000.